\documentclass{article}
\usepackage{cite}
\usepackage{amsmath,amssymb,amsfonts}
\usepackage{algorithmic}
\usepackage{graphicx}
\usepackage{textcomp}
\usepackage{float}
\usepackage{array,multirow,graphicx}
\usepackage{animate}
\usepackage{arydshln}
\usepackage{mathtools}
\usepackage{xmpmulti}
\usepackage{alltt}
\usepackage{bm}
\usepackage{easybmat}
\usepackage{soul}
\usepackage[ruled, vlined]{algorithm2e}
\usepackage[thinlines]{easytable}
\newcolumntype{M}[1]{>{\centering\arraybackslash}m{#1}}
\usepackage{soul}
\usepackage{spconf}

\usepackage[dvipsnames]{xcolor}
\definecolor{ForestGreen}{RGB}{11, 148, 43}
\definecolor{magenta}{RGB}{247,29,176}
\definecolor{orange}{RGB}{219,106,72}
\definecolor{skyblue}{RGB}{42,138,153}
\usepackage{hyperref}
\DeclareMathOperator*{\argmin}{arg\,min}
\DeclareMathOperator*{\minimize}{minimize}
\DeclareMathOperator*{\subto}{subject\,to}
\newcommand{\vect}[1]{\boldsymbol{\mathbf{#1}}}

\newcommand{\iid}{\overset{i.i.d.}{\sim}}
\newcommand{\vTheta}{\vect{\Theta}}
\newcommand{\vD}{\mathbf{D}}
\newcommand{\cM}{\mathcal{M}}
\newcommand{\vPsi}{\vect{\Psi}}
\newcommand{\vQ}{\mathbf{Q}}
\newcommand{\vR}{\mathbf{R}}

\newcommand{\vdelta}{\vect{\delta}}
\newcommand{\vU}{\mathbf{U}}
\newcommand{\vz}{\mathbf{z}}
\newcommand{\vt}{\text{vec}}

\newtheorem{prop}{Proposition}
\begin{document}

\title{Clustered Gaussian Graphical Model via Symmetric Convex Clustering}

\name{$\text{Tianyi Yao}^{a}$ and $\text{Genevera I. Allen}^{b,c}$ \thanks{GA and TY acknowledge support from NSF DMS-1554821 and NSF NeuroNex-1707400.}}
\address{$^a$Dept. of Statistics, Rice University \\
$^b$Dept. of Statistics, Computer Science, and Electrical and Computer Engineering, Rice University \\
$^c$Neurological Research Institute, Baylor College of Medicine \\
Houston, TX}

\maketitle

\begin{abstract}
Knowledge of functional groupings of neurons can shed light on structures of neural circuits and is valuable in many types of neuroimaging studies. However, accurately determining which neurons carry out similar neurological tasks via controlled experiments is both labor-intensive and prohibitively expensive on a large scale. Thus, it is of great interest to cluster neurons that have similar connectivity profiles into functionally coherent groups in a data-driven manner. In this work, we propose the clustered Gaussian graphical model (GGM) and a novel symmetric convex clustering penalty in an unified convex optimization framework for inferring functional clusters among neurons from neural activity data. A parallelizable multi-block Alternating Direction Method of Multipliers (ADMM) algorithm is used to solve the corresponding convex optimization problem. In addition, we establish convergence guarantees for the proposed ADMM algorithm. Experimental results on both synthetic data and real-world neuroscientific data demonstrate the effectiveness of our approach. 
\end{abstract}


\section{INTRODUCTION}
In neuroscience, an important goal is to identify which neurons are involved in similar computations and how they are organized into functionally coherent units to carry out specific computational tasks in the brain. Such knowledge of functional organizations of neurons could lead to a better understanding of structures of interconnected neural circuits and thus the operating mechanisms of the brain. Advancement of optical imaging technologies such as calcium imaging has enabled indirect recordings of spiking activity from thousands of neurons simultaneously \cite{calc1,calc2}. Learning the functional organizations of large neuronal populations from such high-dimensional neural activity recording data is a major challenge in computational neuroscience. 

Functional connectivity, which is defined as statistical dependence among measurements of neuronal activity in \cite{MAGRANSDEABRIL2018120}, has been widely used to describe functional interactions among measured neuronal populations. Because functional connectivity is not directly observable, numerous techniques such as correlations and partial correlations have been proposed to estimate such functional connectivity from neural recording data (see \cite{MAGRANSDEABRIL2018120} for a comprehensive review). In this work, we define functional connectivity between each pair of recorded neurons to be their pairwise partial correlation or edges in an undirected GGM in high dimensions. Because the pairwise partial correlation between two neurons takes activities of all the other recorded neurons into account, it captures only direct associations between neurons and discard all indirect associations \cite{MAGRANSDEABRIL2018120,Sutera2014SimpleCI}, which makes pairwise partial correlation coefficient a better indicator of functional connectivity than Pearson correlation. Furthermore, because pairwise partial correlation is the same as the corresponding off-diagonal entries of the standardized precision matrix, the functional connectivity graph of all recorded neurons can be represented by the standardized precision matrix or the corresponding undirected GGM \cite{lauritzen1996graphical}. 

While there is no standardized definition for functional cluster, many neuroscientific studies have found that each neuronal type has its own distinct input-output connectivity patterns \cite{Jiangaac9462} and neurons with similar connectivity patterns typically have similar neurological roles and functions \cite{Bullmore}. Therefore, in this work, we seek to define functional clusters to be groups of neurons that share functional connectivity patterns. Hence, inferring functionally coherent groups of neurons is equivalent to clustering neurons with similar functional connectivity patterns.

While many techniques have been proposed for uncovering clusters from multivariate data (see \cite{Jain:1999:DCR:331499.331504} for a comprehensive review) as well as for finding community structures in network data (see \cite{Harenberg} for a comprehensive review), they are somewhat limited in this application for various reasons. First of all, distance-based clustering techniques such as k-means and hierarchical clustering on pairwise Euclidean distances cluster variables based on the first-moment of the distribution, whereas functional clusters are defined by functional connectivity patterns, which are characterized by the second-moment of the distribution. Some studies in fMRI have applied hierarchical clustering on empirical partial correlation based dissimilarity matrix to cluster brain regions \cite{Salvador}. However, such approaches are not applicable to high-dimensional neural activity data because the MLE of partial correlation matrix does not exist due to singularity of the empirical covariance matrix. Others have taken a two-step approach where a functional connectivity graph is first estimated and then community detection algorithms are used subsequently to infer clusters \cite{Wednesday2017DAPFV, guo2010modularized}. Yet such two-step approaches are highly sensitive to noise as a single erroneously estimated functional connection in the first step could adversely impact the clustering results of the community detection algorithms. Last but not least, some studies have proposed nonparametric Bayesian approaches for estimating the block structures of GGM and clustering variables using a MCMC sampling method \cite{Sun2014AdaptiveVC}. However, such MCMC-based approaches can easily become computationally infeasible on moderate-sized graphs. 

In this paper, we make several methodological contributions: (1) We propose the clustered GGM that involves a novel symmetric convex clustering penalty, which allows us to exploit the symmetric structures of a functional connectivity graph for better estimation of functional clusters. (2) We provide a tractable ADMM algorithm with convergence guarantees to fit our clustered GGM method in big-data settings. Because of these contributions, our clustered GGM method enjoys many advantages over existing approaches to inferring functional clusters from neural activity data: (i) With our novel symmetric convex clustering penalty, our method explicitly leverages functional connectivity patterns to cluster neurons into functionally coherent groups. (ii) Because the clustered GGM is formulated in an unified convex optimization framework, our single-step method is more stable and conducive to data-driven model selection. 
\vspace{-8pt}
\section{The Clustered GGM}
\subsection{Model Setup and Background}
Suppose the neural activity recordings of $p$ neurons over $n$ time points are arranged into the data matrix $\mathbf{X}\in \mathbb{R}^{n\times p}$ and the recording of all $p$ neurons at the $i$th time point, $\mathbf{X}_i=\{X_{i1},\hdots,X_{ip}\}$, is a random $p$-vector independently drawn from the same time-invariant $p$-variate Gaussian distribution $\mathcal{N}(\mathbf{0}_p,\vect{\Sigma}_{p\times p})$ \cite{Sutera2014SimpleCI}. We can approximately achieve the assumption of independence by prewhitening the raw time series using appropriate time series models. As noted before, the functional connectivity graph can be represented by the standardized precision matrix $\vect{\Theta} \succ 0$, where $\vect{\Theta}_{ij}=-\vect{\Sigma}^{-1}_{ij}/\sqrt{\vect{\Sigma}^{-1}_{ii}\vect{\Sigma}^{-1}_{jj}}$. Hence, estimating functional clusters based on functional connectivity patterns is equivalent to recovering the group structures that form checkerboard patterns in $\vect{\Theta}$.

\subsection{The Symmetric Convex Clustering Penalty}
At first glance, designing a penalty function to encourage checkerboard patterns in the estimate of $\vect{\Theta}$ seems straightforward as one might ask whether we can simply apply the convex biclustering fusion penalty \cite{chi} to simultaneously force rows and columns of $\vect{\Theta}$ to coalesce to form block structures. However, simply applying such biclustering penalty does not guarantee the same amount of fusion along the rows and columns of $\vect{\Theta}$ and it can easily result in different estimated functional clusters between rows and columns. In fact, any fusion penalty that directly regularizes elements of $\vect{\Theta}$ would lead to asymmetric estimates, thus leading to difficult interpretations. Also recognized by \cite{Sun2014AdaptiveVC}, designing a penalty function to force such checkerboard patterns in a GGM in a computationally feasible way is indeed a challenging task.

Our objective is to develop a convex penalty function that allows us to explicitly model functional clusters among neurons based on mutual pairwise functional connectivity patterns and preserve the symmetry of estimated functional connectivity graph as well as neuron cluster assignments. To this end, we build upon the convex fusion penalty \cite{Eric, Hocking:2011:CAC:3104482.3104576} and introduce a novel symmetric convex clustering penalty that encourages symmetric checkerboard patterns in the estimated precision matrix. 

Consider a $p\times p$ symmetric matrix $\vect{\Theta}$, the symmetric convex clustering penalty function takes the form
    \begin{align*}
    P(\vect{\Theta}) = &\sum_{l \in \mathcal{M}} w_l ||\:\vect{\Psi}_{l,1} - \vect{\Psi}_{l,2}\:||_2 \\
    \subto & \:\:(\mathbf{Q}_l\vect{\Theta}\mathbf{R}_l) - \vect{\Psi}_l = 0, \forall l \in \mathcal{M}
    \end{align*}
Here, we index a neuron pair by $l=(i,j)$ with $1\leq i<j\leq p$ and define the fusion set over the non-zero fusion weights $\mathcal{M}=\{l=(i,j)\::\:w_l > 0\}$. The set of all nonnegative, pairwise fusion weights $\{w_l\}_{1\leq i<j\leq p}$ can be specified beforehand to incorporate domain knowledge and take auxiliary information (e.g. interneuron distances) into account. Additionally, $\vect{\Psi}_{l,1}$ (and $\vect{\Psi}_{l,2}$) denotes the $1$st (and $2$nd) column of $\vect{\Psi}_l \in \mathbb{R}^{(p-2)\times 2}$, which can be interpreted as cluster centroid matrix corresponding to the $l=(i,j)$th neuron pair. For $l=(i,j)$, the rows of $\mathbf{Q}_l \in \mathbb{R}^{(p-2)\times p}$ consist of canonical basis vectors $\mathbf{e}_q$ for $q \in \{1,2,\hdots,p\} \setminus \{i,j\}$ and the columns of $\mathbf{R}_l \in \mathbb{R}^{p\times 2}$ consist of canonical basis vectors $\mathbf{e}_r$ for $r \in \{i,j\}$. 

We now discuss the intuition behind the symmetric convex clustering penalty $P(\vect{\Theta})$. For any neuron pair $l=(i,j)$ in the fusion set $\mathcal{M}$, the canonical basis matrices $\mathbf{Q}_l$ and $\mathbf{R}_l$ extracts a portion of $\vect{\Theta}$ such that the $1$st (and $2$nd) column of $\mathbf{Q}_l\vect{\Theta}\mathbf{R}_l \in \mathbb{R}^{(p-2)\times 2}$ represents the functional connectivity patterns of neuron $i$ (and neuron $j$) with all the other recorded neurons. $\vect{\Psi}_l$ is taken to be a copy of $\mathbf{Q}_l\vect{\Theta}\mathbf{R}_l$ and the fusion penalty term $||\:\vect{\Psi}_{l,1} - \vect{\Psi}_{l,2}\:||_2$ induces sparsity in the difference between neuron $i$ and $j$'s respective functional connectivity patterns with all the other recorded neurons, thus encouraging the estimates of $\vect{\Psi}_{l,1}$ and $\vect{\Psi}_{l,2}$ to fuse. Neuron $i$ and $j$ are assigned to the same functional cluster if $\hat{\vect{\Psi}}_{l,1} = \hat{\vect{\Psi}}_{l,2}$, which means neuron $i$ and $j$ have the same conditional relationships with all the other recorded neurons. All such fusions can be done separately and in parallel for each neuron pair $l \in \mathcal{M}$, and the set of equality constraints $\mathbf{Q}_l\vect{\Theta}\mathbf{R}_l - \vect{\Psi}_l = 0, \forall l \in \mathcal{M}$ aggregates all fusion results back to $\vect{\Theta}$ to form symmetric checkerboard patterns denoting functional clusters among neurons. 

\subsection{The Clustered GGM via Symmetric Convex Clustering}
While one can apply the symmetric convex clustering penalty to any loss functions that take a symmetric matrix as input, we specifically apply $P(\vect{\Theta})$ to the negative log-likelihood of the multivariate Gaussian distribution to yield the clustered GGM problem.
    \begin{align*}\label{eq:1}
    \hspace{-1.0cm}\minimize_{\vect{\Theta}\in \mathbb{S}_{++}^p, \{\vect{\Psi}_l\}} &-\log{\text{det}\vect{\Theta}} + \text{trace}(\hat{\vect{\Sigma}}\vect{\Theta}) \\
    &+ \lambda\sum_{l \in \mathcal{M}} w_l ||\:\vect{\Psi}_{l,1} - \vect{\Psi}_{l,2}\:||_2\tag{1} \\
    \subto &\:\:(\mathbf{Q}_l\vect{\Theta}\mathbf{R}_l) - \vect{\Psi}_l = 0, \forall l \in \mathcal{M}
    \end{align*}
where $\mathbb{S}_{++}^p$ denotes the set of positive definite matrices of size $p$ and $\hat{\vect{\Sigma}} = \frac{1}{n}\mathbf{X}^T\mathbf{X}$ denotes the empirical covariance matrix (assuming the columns of $\mathbf{X}$ are properly centered). Unlike the GLasso problem \cite{Yuan}, our clustered GGM does not aim to produce a sparse graph estimate. Instead, our clustered GGM leads to a graph estimate $\hat{\vect{\Theta}}$ with block structures that indicate cluster assignments of nodes. In addition, fusing many elements of $\vect{\Theta}$ to the same values, the symmetric convex clustering penalty significantly reduces the effective number of parameters to be estimated, thus making our clustered GGM an attractive choice for high-dimensional settings. The amount of fusion, and hence the number of clusters, is determined by the penalty parameter $\lambda$. The optimal $\lambda$ can be chosen via data-driven model selection techniques such as consensus clustering \cite{Monti2003}.

\subsection{The Clustered GGM Algorithm}
We adopt the generalized ADMM framework described in \cite{Boyd:2011:DOS:2185815.2185816,Deng2017} as well as an approach introduced in \cite{Eric} for convex clustering problems in order to develop a tractable 3-block ADMM algorithm to solve (\ref{eq:1}). ADMM is an appealing algorithm for this problem because it permits us to decouple the terms in (\ref{eq:1}) that are challenging to jointly optimize. Specifically, we reformulate (\ref{eq:1}) by introducing a set of auxiliary variables $\{\vect{\delta}_l\}$ and rewrite the penalty term in terms of these auxiliary variables. \\
\begin{align*}\label{eq:2}
\minimize_{\vect{\Theta}\in\mathbb{S}_{++}^p, \{\vect{\Psi}_l\}, \{\vect{\delta}_l\}} &-\log{\text{det}\vect{\Theta}} + \text{trace}(\hat{\vect{\Sigma}}\vect{\Theta}) \\
&+ \lambda\sum_{l \in \mathcal{M}} w_l ||\:\vect{\delta}_l\:||_2\tag{2} \\
\subto \:\:\:&\mathbf{Q}_l\vect{\Theta}\mathbf{R}_l - \vect{\Psi}_l = 0, \forall l \in \mathcal{M} \\
\:\:\:&\vect{\Psi}_{l,1} - \vect{\Psi}_{l,2} - \vect{\delta}_l = 0, \forall l \in \mathcal{M}
\end{align*}
Following from \cite{Boyd:2011:DOS:2185815.2185816,Eric,Deng2017}, we give Algorithm \ref{alg:1} to solve the clustered GGM problem:
 
$\nabla_{\vect{\Theta}}\mathcal{L}(\vect{\Theta}^{(k,j-1)})$ denotes the gradient of the corresponding augmented Lagrangian in scaled form evaluated at $\vect{\Theta}^{(k,j-1)}$ and $s_{j}$ is the stepsize of gradient descent, which can be selected via the Goldstein-Armijo line search procedure. Here, $k$ is the iteration counter for the outer 3-block ADMM updates and $j$ is the iteration counter for the inner gradient descent updates for the $\vect{\Theta}$ subproblem. $\mathbf{e}_1, \mathbf{e}_2 \in \mathbb{R}^{2}$ are canonical basis vectors and $\mathbf{D}\in\mathbb{R}^{(p-2)\times 2(p-2)}=(\mathbf{e}_1-\mathbf{e}_2)^T \otimes \mathbf{I}_{(p-2)}$ is the directed difference matrix such that $\mathbf{D}\text{vec}(\vect{\Psi}_l) = \vect{\Psi}_{l,1} - \vect{\Psi}_{l,2}$. Convergence of the algorithm is measured by the norm of the primal and dual residuals and the parameters $\rho_1, \rho_2 > 0$ are fixed throughout the algorithm as recommended by \cite{Boyd:2011:DOS:2185815.2185816}. 

\begin{algorithm}\label{alg:1}
 \caption{ADMM algorithm for the clustered GGM}
\SetAlgoLined
\KwIn{$\hat{\vect{\Sigma}}$, $\lambda \geq 0$, $\rho_1, \rho_2 > 0$}
\vspace{5pt}
 {\bf Initialize:} Primal variables to identity matrices and dual variables to zero matrices\;
 \vspace{5pt}
 {\bf Precompute:} $\mathbf{D}$, $\{w_l\}$, $\mathcal{M}$\;
 \vspace{5pt}

 \While{not converged}{
 \vspace{5pt}
 (i) Update $\vect{\Theta}$:\\
 \vspace{5pt}
  \While{not converged}{
\vspace{5pt}
$\vect{\Theta}^{(k,j)} \leftarrow \vect{\Theta}^{(k,j-1)} - s_{j} \nabla_{\vect{\Theta}}\mathcal{L}(\vect{\Theta}^{(k,j-1)})$\;
\vspace{5pt}
}
\vspace{5pt}
(ii) Update $\vect{\Psi}_l$ ($\forall l \in \mathcal{M}$ in parallel): \\
\vspace{5pt}
\-\hspace{0.5cm}$\vect{\Psi}_l^{(k)} \leftarrow \frac{\rho_1}{\rho_1+2\rho_2}[\mathbf{Q}_l\vect{\Theta}^{(k)}\mathbf{R}_l + \mathbf{U}_l^{(k-1)} +$ 
\vspace{5pt}
\hspace{0.6cm}$\frac{\rho_2}{\rho_1}(\vect{\delta}_l^{(k-1)}-\mathbf{z}_l^{(k-1)})(\mathbf{e}_1-\mathbf{e}_2)^T](\mathbf{I}_2 + \frac{\rho_2}{\rho_1}\mathbf{1}\mathbf{1}^T)$\;
\vspace{5pt}
(iii) Update $\vect{\delta}_l$ ($\forall l \in \mathcal{M}$ in parallel): \\
\vspace{5pt}
\-\hspace{0.5cm} $\vect{\delta}_{l}^{(k)} \leftarrow \text{prox}_{\frac{\lambda w_l}{\rho_2}||.||_{2}}(\mathbf{D}\text{vec}(\vect{\Psi}_{l}^{(k)}) + \mathbf{z}_{l}^{(k-1)})$\;
\vspace{5pt}
(iv) Update $\mathbf{U}_l$ ($\forall l \in \mathcal{M}$ in parallel): \\
\vspace{5pt}
\-\hspace{0.5cm} $\mathbf{U}_l^{(k)} \leftarrow \mathbf{U}_l^{(k-1)} + (\mathbf{Q}_l\vect{\Theta}^{(k)}\mathbf{R}_l - \vect{\Psi}_l^{(k)})$\;
\vspace{5pt}
(v) Update $\mathbf{z}_l$ ($\forall l \in \mathcal{M}$ in parallel): \\
\vspace{5pt}
\-\hspace{0.5cm} $\mathbf{z}_l^{(k)} \leftarrow \mathbf{z}_{l}^{(k-1)} + (\mathbf{D}\text{vec}(\vect{\Psi}_{l}^{(k)}) - \vect{\delta}_{l}^{(k)})$\;
}
\end{algorithm}

\begin{prop}
Algorithm \ref{alg:1} converges to a global solution to problem (\ref{eq:1}).
\end{prop}
{\it Proof sketch:} We can first recast our 3-block ADMM problem (\ref{eq:2}) as the general 3-block ADMM formulation described in \cite{Chen2016} by re-writing the set of equality constraints in (\ref{eq:2}) as a linear combination of the three optimization variables:

$\mathbf{A}_1\text{vec}(\vect{\Theta}) + 
\mathbf{A}_2 \text{vec}(\vect{\Psi}) + 
\mathbf{A}_3 \text{vec}(\vect{\Delta}) = \mathbf{0}$, \\
where $\vect{\Psi}=[\vect{\Psi}_1,\hdots,\vect{\Psi}_{|\mathcal{M}|}]$, $\vect{\Delta}=[\vect{\delta}_1,\hdots,\vect{\delta}_{|\mathcal{M}|}]$,\\ $\mathbf{A}_1=
\left[
\begin{array}{c}
\mathbf{B}_{2g\times p^2}  \\
\mathbf{0}_{g\times p^2}
\end{array}
\right]$,
$\mathbf{A}_2 = 
\left[
\begin{array}{c}
-\mathbf{I}_{2g}  \\
\mathbf{H}_{g \times 2g}
\end{array}
\right]$,
$\mathbf{A}_3 = 
\left[
\begin{array}{c}
\mathbf{0}_{2g\times g}  \\
-\mathbf{I}_{g}
\end{array}
\right]$ \\
with rows of $\mathbf{B}$ containing appropriate canonical basis vectors and $\mathbf{H}$ containing $|\mathcal{M}|$ directed difference matrices $\mathbf{D}$ on its diagonal. For notational simplicity, we use $g = (p-2)|\mathcal{M}|$. With $\mathbf{A}_1^T\mathbf{A}_3 = \mathbf{0}$, we can show that (\ref{eq:2}) satisfies the sufficient conditions (Theorem 2.4 in \cite{Chen2016}) for the convergence of such 3-block ADMM algorithm.

\section{Experiments}
\subsection{Synthetic Data}
In this subsection, we evaluate the comparative performance of our clustered GGM method on simulated data sets.
\subsubsection{Data Generation}

Suppose we have $p$ neurons which form $k$ functional clusters, we simulate a standardized precision matrix $\vect{\Theta}$ with the desired checkerboard patterns reflecting groundtruth functional clusters as follows: first we define the groundtruth cluster membership for the $p$ neurons by creating $\mathbf{Z}_{p\times k}\in\{0,1\}$ which has exactly one $1$ in each row and at least one $1$ in each column. We then generate symmetric matrix $\mathbf{B}_{k\times k}\in[-1,1]$ where $B_{ii}$ denotes the partial correlation between two neurons if both neurons are in the $i$th cluster and $B_{ij}$ denotes the partial correlation between a neuron from the $i$th cluster and a neuron from the $j$th cluster. Specifically, $B_{ii} \iid \text{Unif}([0.6,0.95])$ and $B_{ij} \iid \text{Unif}([0,0.55])$. Next, we generate the groundtruth precision matrix $\vect{\Theta} = \mathbf{Z}\mathbf{B}\mathbf{Z}^T$ and set the diagonal entries of $\vect{\Theta}$ to $1$'s to ensure positive-definiteness. Finally, we generate the data matrix $\mathbf{X} \in \mathbb{R}^{n\times p}$ according to $\mathbf{x}_1,\hdots,\mathbf{x}_n \iid \mathcal{N}(\mathbf{0},\vect{\Theta}^{-1})$. We consider two simulation scenarios: Scenario I with $n=110$, and $p=50$ neurons are randomly divided into $k=3$ clusters with size $5$, $15$, and $30$, respectively; Scenario II with $n=200$, and $p=200$ neurons are randomly divided into $k=3$ clusters with size $30$, $60$, and $110$, respectively. 

\subsubsection{Results}

We compare our clustered GGM to other popular clustering approaches: 1) k-means; 2) Hierarchical Clustering (HC) with various linkage functions and dissimilarity metrics (Euclidean distance and empirical correlation); 3) Spectral Clustering (SC) with various similarity metrics (empirical correlation and various kernel functions), implemented using {\tt R} packages {\tt anocva} and {\tt kernlab}; 4) GLasso followed by commonly used community detection algorithms such as the Louvain method \cite{Blondel_2008}, implemented using {\tt R} packages {\tt huge} and {\tt igraph}. The best penalty parameter for the GLasso is selected by the {\tt ebic} criterion embedded in {\tt huge}. Moreover, the oracle number of functional clusters $k=3$ is supplied to all aforementioned clustering techniques. Such practice is reasonable in the neuroscientific context because the number of functional clusters is typically known {\it a priori} from domain knowledge. 

In Table \ref{table_1}, results on functional cluster recovery are presented. In particular, the performance in terms of functional cluster recovery is quantified using Rand Index which measures the agreement between the unsupervised clustering solutions and the true cluster membership. Rand Index takes values between $0$ and $1$ with 1 indicating perfect cluster recovery. We only include the best performing approaches from each category 2), 3), and 4) in Table \ref{table_1}. Results in Table \ref{table_1} reveal that our clustered GGM outperforms all competing approaches in terms of functional cluster recovery.

\begin{table}[!t]
\renewcommand{\arraystretch}{1.3}
\caption{Simulation results averaged over $10$ replicates in terms of Rand Index. Best performing methods are boldfaced.}
\label{table_1}
\centering
\begin{tabular}{|c|c|c|}
\hline
Dataset & Method & Rand Index\\
\hline
\multirow{6}{4em}{Scenario I} & Clustered GGM & \bfseries 0.964 (0.068) \\
& k-means & 0.505 (0.003) \\
& HC Euclidean Ward & 0.498 (0.007) \\
& SC empirical corr & 0.511 (0.006) \\
& HC empirical corr Ward & 0.521 (0.027) \\
& GLasso $+$ Louvain & 0.556 (0.022) \\
\hline
\multirow{6}{4em}{Scenario II} & Clustered GGM & \bfseries 0.999 (0.003) \\
& k-means & 0.515 (0.01) \\
& HC Euclidean Ward & 0.791 (0.003) \\
& SC empirical corr & 0.526 (0.002) \\
& HC empirical corr Ward & 0.513 (0.023) \\
& GLasso $+$ Louvain & 0.566 (0.003) \\
\hline
\end{tabular}
\end{table}

\subsection{Case Study: Calcium Imaging Data}
\begin{figure*}[t]
  \centering
    \includegraphics[width=0.35\textwidth]{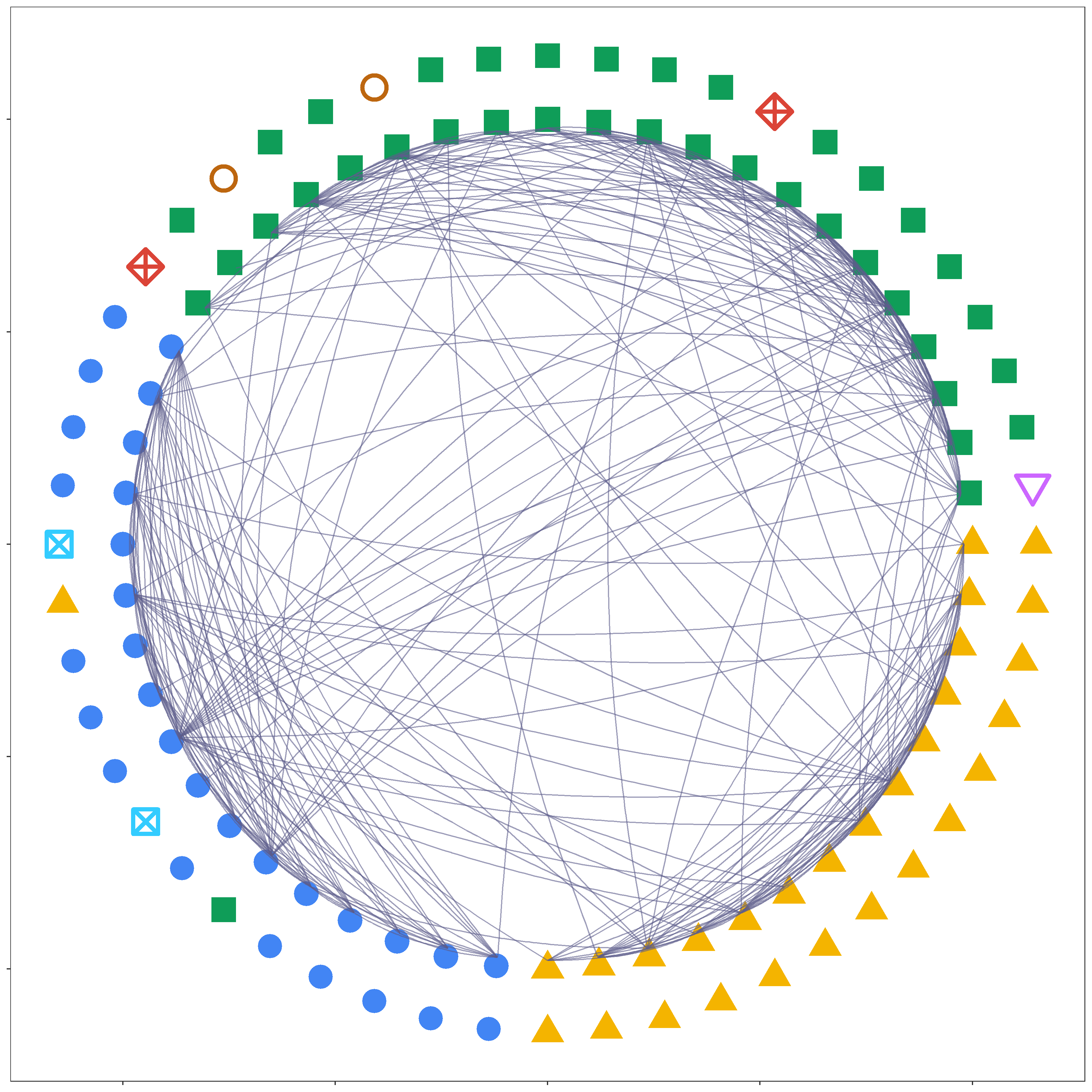}
\caption{Comparison of empirically determined neuron tuning labels (inner circle) and functional cluster labels estimated by the clustered GGM (outer circle). Nodes on the inner circle are colored according to neuron tuning labels whereas nodes on the outer circle are colored according to functional cluster labels estimated by the clustered GGM. The Rand Index between neuron tuning labels and functional cluster labels estimated by the clustered GGM is $0.868$.}
\label{fig:2}
\end{figure*}
We test our method on a publicly available calcium imaging data set from \cite{Stringer306019,Stringer2018}. Neural activity of a subset of excitatory neurons in mouse visual cortex was recorded using multi-plane acquisition and $10$ to $12$ planes of different depth were recorded at the same time at a sampling rate of about $3$Hz (see \cite{Stringer306019} for detailed data acquisition and processing procedures). During the course of experiments, $32$ natural images were shown to an awake mouse sequentially and averaged responses of each recorded neuron to the visual stimuli were determined after adjusting for trial-to-trial variability via model-based approaches. Each neuron is said to be tuned to the natural image to which it had the largest averaged responses and was subsequently assigned a neuron tuning label. Such neuron tuning labels are often used as estimates of functional clusters. However, such empirically inferred neuron tuning labels are likely to be noisy and there could be considerable amount of uncertainty associated with functional groups determined solely by such tuning labels. In this case study, we seek to evaluate how well the noisy neuron tuning labels serve as proxies for identifying functional clusters of neurons in mouse visual cortex. 

We select $52$ excitatory neurons residing in the most superficial imaging plane that were empirically determined to be tuned to three most dissimilar natural images. The calcium imaging data come in the form of deconvolved calcium traces, whose distributions are highly skewed. To accommodate our model assumptions of independence and Gaussianity, we prewhiten individual calcium traces with an autoregressive model of order $1$ to remove temporal dependence and subsequently perform the semiparametric copula transformation \cite{Liu:2009:NSE:1577069.1755863} to make the data approximately follow a multivariate Gaussian distribution. Afterwards, we apply our clustered GGM to the processed traces of these $52$ neurons across 855 time points at stimulus onset. Specifically, we fit the clustered GGM to the data on a fine grid of penalty parameter values $\lambda \in [0,1.92]$ such that all neurons are clustered into one group for $\lambda \geq 1.92$. The best penalty parameter value selected is $\lambda = 1.06$ and the corresponding estimated functional clusters are displayed in the right panel of Fig. \ref{fig:2}.

In Fig. \ref{fig:2}, each node denotes a neuron and edges represent the functional connectivity graph. Nodes on the inner circle are colored according to the noisy neuron tuning labels whereas nodes on the outer circle are colored based upon estimated functional cluster labels by our clustered GGM. The Rand Index between neuron tuning labels and functional cluster labels estimated by our clustered GGM is $0.868$. Such results show that the functional clusters estimated by our clustered GGM largely agree with the empirically determined neuron tuning labels except for a handful of singletons, suggesting that neuron tuning labels serve as good proxies for identifying functional clusters of neurons in mouse visual cortex.

\section{CONCLUSIONS}
In this paper, we have introduced the clustered GGM via symmetric convex clustering in an unified convex optimization framework, which can be used to infer functional clusters among neurons from neural activity recordings. Key contributions include developing a novel symmetric convex clustering penalty to explicitly group neurons with similar functional connectivity patterns as well as providing a tractable algorithm to solve the clustered GGM problem with notable convergence guarantees. Experimental results on both synthetic data and real-world neuroscientific data demonstrate the effectiveness of our proposed method.

Even though the focus of this paper has been on the clustered GGM problem, our novel symmetric convex clustering penalty can be applied to many other convex loss functions that take symmetric matrices as inputs. Such flexibility of our novel penalty function suggests that there is potential for broad application of our approach to data in areas such as genomics and proteomics. 

\bibliographystyle{IEEEbib}
\bibliography{./mybib11}

\newpage
\onecolumn
\section{Detailed Derivations}
In this section, we provide derivations of the ADMM algorithm in Algorithm \ref{alg:1} for the clustered GGM. Our notation here is the same as that used in the main body of the paper unless otherwise stated. \\
After introducing the set of auxiliary variables $\{\vdelta_l\}$ and rewriting the clustered GGM problem as (\ref{eq:2}), the augmented Lagrangian in scaled form is given by:
\begin{align*}
    \mathcal{L}_{\rho_1,\rho_2}(\vTheta,\{\vPsi_l\},\{\vdelta_l\},\{\vU_l\},\{\vz_l\}) &= -\log{\text{det}\vect{\Theta}} + \text{trace}(\hat{\vect{\Sigma}}\vect{\Theta}) + \lambda\sum_{l \in \mathcal{M}} w_l ||\:\vect{\delta}_l\:||_2 \\
    &+ \frac{\rho_1}{2}\sum_{l\in\cM} \Big(||\mathbf{Q}_l\vect{\Theta}\mathbf{R}_l - \vect{\Psi}_l + \vU_l||_F^2 - ||\vU_l||_F^2\Big) \\
    &+ \frac{\rho_2}{2}\sum_{l\in\cM} \Big(||\vD \vt(\vPsi_l) - \vect{\delta}_l + \vz_l||_2^2 - ||\vz_l||_2^2\Big)
\end{align*}
Following from \cite{Boyd:2011:DOS:2185815.2185816}, the scaled form of the ADMM updates are given by:
\begin{align*}
    \vTheta^{(k)} &= \argmin_{\vTheta \in \mathbb{S}^p_{++}} -\log{\text{det}\vect{\Theta}} + \text{trace}(\hat{\vect{\Sigma}}\vect{\Theta}) + \frac{\rho_1}{2}\sum_{l\in\cM} ||\mathbf{Q}_l\vect{\Theta}\mathbf{R}_l - \vect{\Psi}_l^{(k-1)} + \vU_l^{(k-1)}||_F^2 \\
    \vPsi_l^{(k)} &= \argmin_{\vPsi_l\in\mathbb{R}^{(p-2)\times 2}}\frac{\rho_1}{2}||\mathbf{Q}_l\vect{\Theta}^{(k)}\mathbf{R}_l - \vect{\Psi}_l + \vU_l^{(k-1)}||_F^2 + \frac{\rho_2}{2}||\vD \vt(\vPsi_l) - \vect{\delta}_l^{(k-1)} + \vz_l^{(k-1)}||_2^2, \:\:\forall l \in \cM \\
    \vdelta_l^{(k)} &= \argmin_{\vdelta_l \in \mathbb{R}^{p-2}} \lambda w_l ||\:\vect{\delta}_l\:||_2 + \frac{\rho_2}{2}||\vD \vt(\vPsi_l^{(k)}) - \vect{\delta}_l + \vz_l^{(k-1)}||_2^2, \:\: \forall l \in \cM \\
    \vU_l^{(k)} &= \vU_l^{(k-1)}  + \mathbf{Q}_l\vect{\Theta}^{(k)}\mathbf{R}_l - \vect{\Psi}_l^{(k)}, \:\: \forall l \in \cM \\
    \vz_l^{(k)} &= \vz_l^{(k-1)} + \vD \vt(\vPsi_l) - \vect{\delta}_l, \:\: \forall l \in \cM
\end{align*}
where $\{\vU_l\}_{l\in\cM}$ and $\{\vz_l\}_{l\in\cM}$ are the corresponding dual variables. First, we consider the $\vTheta$-update:
\begin{align*}
    \vTheta^{(k)} = \argmin_{\vTheta \in \mathbb{S}^p_{++}} -\log{\text{det}\vect{\Theta}} + \text{trace}(\hat{\vect{\Sigma}}\vect{\Theta}) + \frac{\rho_1}{2}\sum_{l\in\cM} ||\mathbf{Q}_l\vect{\Theta}\mathbf{R}_l - \vect{\Psi}_l^{(k-1)} + \vU_l^{(k-1)}||_F^2
\end{align*}
This is smooth and so we compute the gradient with respect to $\vTheta$:
\begin{align*}
    \nabla_{\vTheta}\mathcal{L} = -\vTheta^{-1} + \hat{\vect{\Sigma}} + \frac{\rho_1}{2} \sum_{l\in\cM} (2\vQ_l^T\vQ_l \vTheta \vR_l\vR_l^T - 2\vQ_l^T \vPsi_l \vR_l^T + 2\vQ_l^T \vU_l \vR_l^T)
\end{align*}
using the identity \footnote{See Equation (119) in the Matrix Cookbook: \href{https://www.math.uwaterloo.ca/~hwolkowi/matrixcookbook.pdf}{https://www.math.uwaterloo.ca/~hwolkowi/matrixcookbook.pdf}}
\begin{align*}
    \frac{\partial}{\partial \vTheta}||\mathbf{A}\vTheta\mathbf{B} + \mathbf{C}||_F^2 = 2\mathbf{A}^T (\mathbf{A}\vTheta \mathbf{B} + \mathbf{C})\mathbf{B}^T.
\end{align*}
Then the solution $\vTheta^{(k)}$ to the first subproblem can be obtained by applying gradient descent to convergence.\\
Now we consider the $\vPsi_l$-update:
\begin{align*}
    \vPsi_l^{(k)} &= \argmin_{\vPsi_l\in\mathbb{R}^{(p-2)\times 2}}\frac{\rho_1}{2}||\mathbf{Q}_l\vect{\Theta}^{(k)}\mathbf{R}_l - \vect{\Psi}_l + \vU_l^{(k-1)}||_F^2 + \frac{\rho_2}{2}||\vD \vt(\vPsi_l) - \vect{\delta}_l^{(k-1)} + \vz_l^{(k-1)}||_2^2 \\
    &= \argmin_{\vt(\vPsi_l)\in\mathbb{R}^{2(p-2)}}\frac{\rho_1}{2}||\vt(\mathbf{Q}_l\vect{\Theta}^{(k)}\mathbf{R}_l) - \vt(\vect{\Psi}_l) + \vt(\vU_l^{(k-1)})||_2^2 + \frac{\rho_2}{2}||\vD \vt(\vPsi_l) - \vect{\delta}_l^{(k-1)} + \vz_l^{(k-1)}||_2^2
\end{align*}
Since this is fully smooth, we take the gradient with respect to $\vt(\vPsi_l)$ to obtain the stationarity conditions: 
\begin{align*}
    -\rho_1\Big(\vt(\mathbf{Q}_l\vect{\Theta}^{(k)}\mathbf{R}_l) - \vt(\vect{\Psi}_l) + \vt(\vU_l^{(k-1)})\Big) + \rho_2\mathbf{D}^T\Big(\vD \vt(\vPsi_l) - \vect{\delta}_l^{(k-1)} + \vz_l^{(k-1)}\Big) = \mathbf{0}.
\end{align*}
Re-arranging the terms, we obtain
\begin{align*}\label{eq:3}
    (\rho_1\mathbf{I}_{2(p-2)} + \rho_2 \mathbf{D}^T\mathbf{D})\vt(\vPsi_l) = \rho_1(\vt(\mathbf{Q}_l\vect{\Theta}^{(k)}\mathbf{R}_l) + \vt(\vU_l^{(k-1)})) + \rho_2\mathbf{D}^T(\vect{\delta}_l^{(k-1)} - \vz_l^{(k-1)})\tag{3}
\end{align*}
Though an analytical solution can be obtained by:
\begin{align*}
    \vt(\vPsi_l) = (\rho_1\mathbf{I}_{2(p-2)} + \rho_2 \mathbf{D}^T\mathbf{D})^{-1}[\rho_1(\vt(\mathbf{Q}_l\vect{\Theta}^{(k)}\mathbf{R}_l) + \vt(\vU_l^{(k-1)})) + \rho_2\mathbf{D}^T(\vect{\delta}_l^{(k-1)} - \vz_l^{(k-1)})]
\end{align*}
This update can quickly become computationally expensive as the dimension $p$ grows due to matrix inversion. To avoid such explicit computation of matrix inverse, we exploit the special structure in $(\rho_1\mathbf{I}_{2(p-2)} + \rho_2 \mathbf{D}^T\mathbf{D})$ and take an approach that parallel those of the ADMM for the completely connected convex clustering problem \cite{Eric}. By definition, $\mathbf{D}\in\mathbb{R}^{(p-2)\times 2(p-2)}=(\mathbf{e}_1-\mathbf{e}_2)^T \otimes \mathbf{I}_{(p-2)}$ is the directed difference matrix and $\mathbf{D}^T\mathbf{D}$ can be simplified as follows:
\begin{align*}
    \mathbf{D}^T\mathbf{D} &= [(\mathbf{e}_1-\mathbf{e}_2) \otimes \mathbf{I}_{(p-2)}][(\mathbf{e}_1-\mathbf{e}_2)^T \otimes \mathbf{I}_{(p-2)}] \\
    &= [(\mathbf{e}_1-\mathbf{e}_2)(\mathbf{e}_1-\mathbf{e}_2)^T] \otimes \mathbf{I}_{(p-2)} \\
    &= (2\mathbf{I}_2 - \mathbf{1}\mathbf{1}^T) \otimes \mathbf{I}_{(p-2)}
\end{align*}
using the identity $(\mathbf{A}\otimes \mathbf{B})(\mathbf{C}\otimes \mathbf{D}) = (\mathbf{AC}\otimes \mathbf{BD})$ and $(\mathbf{e}_1-\mathbf{e}_2)(\mathbf{e}_1-\mathbf{e}_2)^T = 2\mathbf{I}_2 - \mathbf{1}\mathbf{1}^T$.
Expanding $(\rho_1\mathbf{I}_{2(p-2)} + \rho_2 \mathbf{D}^T\mathbf{D})$, we obtain:
\begin{align*}
    \rho_1\mathbf{I}_{2(p-2)} + \rho_2 \mathbf{D}^T\mathbf{D} &= \rho_1[\mathbf{I}_2\otimes \mathbf{I}_{p-2}] + \rho_2[(2\mathbf{I}_2 - \mathbf{1}\mathbf{1}^T) \otimes \mathbf{I}_{(p-2)}] \\
    &= [\rho_1\mathbf{I}_2 + \rho_2(2\mathbf{I}_2 - \mathbf{1}\mathbf{1}^T)] \otimes \mathbf{I}_{(p-1)}
\end{align*}
using the identity $\mathbf{A}\otimes\mathbf{C} + \mathbf{B}\otimes \mathbf{C} = (\mathbf{A}+\mathbf{B})\otimes \mathbf{C}$. Now the LHS of (\ref{eq:3}) becomes:
\begin{align*}
    (\rho_1\mathbf{I}_{2(p-2)} + \rho_2 \mathbf{D}^T\mathbf{D})\vt(\vPsi_l) &= \Big([\rho_1\mathbf{I}_2 + \rho_2(2\mathbf{I}_2 - \mathbf{1}\mathbf{1}^T)] \otimes \mathbf{I}_{(p-1)}\Big) \vt(\vPsi_l) \\
    &= \vt\Big(\vPsi_l [\rho_1\mathbf{I}_2 + \rho_2(2\mathbf{I}_2 - \mathbf{1}\mathbf{1}^T)]\Big)
\end{align*}
using the identity $[\mathbf{A}^T\otimes \mathbf{I}]\vt(\mathbf{B}) = \vt(\mathbf{B}\mathbf{A})$. Similarly, the RHS of (\ref{eq:3}) can be re-written as follows:
\begin{align*}
    \text{RHS} &= \rho_1\vt(\mathbf{Q}_l\vect{\Theta}^{(k)}\mathbf{R}_l + \vU_l^{(k-1)}) + \rho_2\vt\Big((\vect{\delta}_l^{(k-1)} - \vz_l^{(k-1)})(\mathbf{e}_1 - \mathbf{e}_2)^T\Big)
\end{align*}
Hence, equation (\ref{eq:3}) can be re-written as 
\begin{align*}\label{eq:4}
    \vt\Big(\vPsi_l [\rho_1\mathbf{I}_2 + \rho_2(2\mathbf{I}_2 - \mathbf{1}\mathbf{1}^T)]\Big) = \rho_1\vt(\mathbf{Q}_l\vect{\Theta}^{(k)}\mathbf{R}_l + \vU_l^{(k-1)}) + \rho_2\vt\Big((\vect{\delta}_l^{(k-1)} - \vz_l^{(k-1)})(\mathbf{e}_1 - \mathbf{e}_2)^T\Big) \tag{4}
\end{align*}
Un-vectorizing both sides of (\ref{eq:4}), we obtain:
\begin{align*}\label{eq:5}
    \vPsi_l [(1+2\frac{\rho_2}{\rho_1})\mathbf{I}_2 - \frac{\rho_2}{\rho_1}\mathbf{1}\mathbf{1}^T] = \mathbf{Q}_l\vect{\Theta}^{(k)}\mathbf{R}_l + \vU_l^{(k-1)} + \frac{\rho_2}{\rho_1}(\vect{\delta}_l^{(k-1)} - \vz_l^{(k-1)})(\mathbf{e}_1 - \mathbf{e}_2)^T \tag{5}
\end{align*}
Applying the Sherman-Morrison formula, we can write the inverse of $[(1+2\frac{\rho_2}{\rho_1})\mathbf{I}_2 - \frac{\rho_2}{\rho_1}\mathbf{1}\mathbf{1}^T]$ as 
\begin{align*}
    [(1+2\frac{\rho_2}{\rho_1})\mathbf{I}_2 - \frac{\rho_2}{\rho_1}\mathbf{1}\mathbf{1}^T]^{-1} = \frac{1}{1+2\frac{\rho_2}{\rho_1}}(\mathbf{I}_2 + \frac{\rho_2}{\rho_1}\mathbf{1}\mathbf{1}^T).
\end{align*}
Therefore, solving (\ref{eq:5}) for $\vPsi_l$, we obtain
\begin{align*}
    \vPsi_l^{(k)} = \frac{1}{1+2\frac{\rho_2}{\rho_1}}[\mathbf{Q}_l\vect{\Theta}^{(k)}\mathbf{R}_l + \vU_l^{(k-1)} + \frac{\rho_2}{\rho_1}(\vect{\delta}_l^{(k-1)} - \vz_l^{(k-1)})(\mathbf{e}_1 - \mathbf{e}_2)^T](\mathbf{I}_2 + \frac{\rho_2}{\rho_1}\mathbf{1}\mathbf{1}^T).
\end{align*}
To solve the third subproblem, we note that it can be written as a proximal operator:
\begin{align*}
   \vdelta_l^{(k)} &= \argmin_{\vdelta_l \in \mathbb{R}^{p-2}} \lambda w_l ||\:\vect{\delta}_l\:||_2 + \frac{\rho_2}{2}||\vD \vt(\vPsi_l^{(k)}) - \vect{\delta}_l + \vz_l^{(k-1)}||_2^2 \\
   &= \argmin_{\vdelta_l \in \mathbb{R}^{p-2}} \frac{\lambda w_l}{\rho_2} ||\:\vect{\delta}_l\:||_2 + \frac{1}{2}||\vect{\delta}_l -(\vD \vt(\vPsi_l^{(k)})+ \vz_l^{(k-1)})||_2^2 \\
   &= \text{prox}_{\frac{\lambda w_l}{\rho_2}||.||_2}(\vD \vt(\vPsi_l^{(k)})+ \vz_l^{(k-1)})
\end{align*}
\end{document}